\pdfoutput=1
\documentclass[11pt]{article}

\usepackage[final]{acl}

\usepackage{times}
\usepackage{latexsym}
\usepackage[T1]{fontenc}
\usepackage[table]{xcolor}
\usepackage[utf8]{inputenc}

\usepackage{microtype}

\usepackage{inconsolata}

\usepackage{graphicx}

\usepackage{siunitx}
\usepackage{algorithm}
\usepackage{algcompatible}
\usepackage{algpseudocode}

\usepackage{bbm}
\usepackage{amsmath}
\usepackage{amssymb}
\usepackage{booktabs}

\usepackage{enumitem}
\usepackage{multirow}

\usepackage{tcolorbox}

\newcommand{\model}{\textbf{II-KEA}}

\title{No Black Boxes: Interpretable and Interactable Predictive Healthcare with Knowledge-Enhanced Agentic Causal Discovery}

\usepackage{authblk} 

\author{
\textbf{Xiaoxue Han},
\textbf{Pengfei Hu},
\textbf{Chang Lu},
\textbf{Jun-En Ding},
\textbf{Feng Liu},
\textbf{Yue Ning}
\\
\normalsize Stevens Institute of Technology \\
\texttt{\{xhan26, phu9, clu13, jding17, fliu22, yue.ning\}@stevens.edu}
}

\begin{document}
\maketitle
\begin{abstract}

Deep learning models trained on extensive Electronic Health Records (EHR) data have achieved high accuracy in diagnosis prediction, offering the potential to assist clinicians in decision-making and treatment planning. However, these models lack two crucial features that clinicians highly value: interpretability and interactivity.
The ``black-box'' nature of these models makes it difficult for clinicians to understand the reasoning behind predictions, limiting their ability to make informed decisions. Additionally, the absence of interactive mechanisms prevents clinicians from incorporating their own knowledge and experience into the decision-making process.
To address these limitations, we propose \model{}, a knowledge-enhanced agent-driven causal discovery framework that integrates personalized knowledge databases and agentic LLMs. \model{} enhances interpretability through explicit reasoning and causal analysis, while also improving interactivity by allowing clinicians to inject their knowledge and experience through customized knowledge bases and prompts.
\model{} is evaluated on both MIMIC-III and MIMIC-IV, demonstrating superior performance along with enhanced interpretability and interactivity, as evidenced by its strong results from extensive case studies. Our code is available at \href{https://github.com/hanxiaoxue114/II-KEA_HealthcareAgent}{https://github.com/hanxiaoxue114/II-KEA\_HealthcareAgent}. 

\end{abstract}

\section{Introduction}
\label{sec:intro}


\begin{figure}[t]
  \centering
  \includegraphics[width=\linewidth]{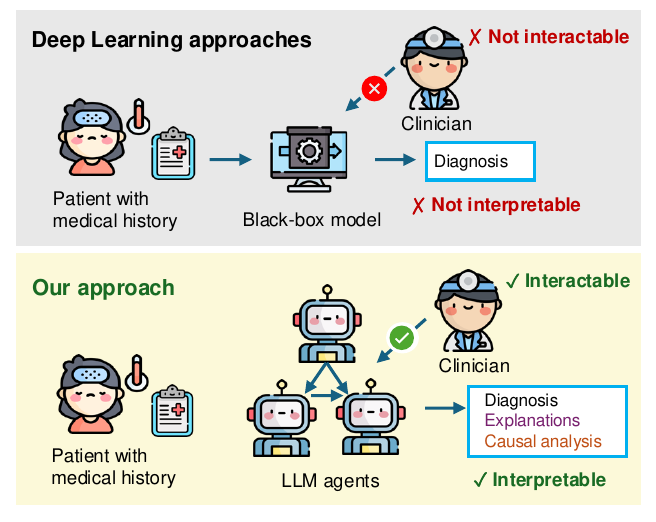}
  \caption{Comparison between deep learning approaches and our approach.}
  \label{fig:nodes}
\end{figure}

Accurate diagnosis prediction is crucial for improving clinical outcomes by enabling timely interventions and optimizing treatment planning. In recent years, the growing availability of Electronic Health Records (EHR) (e.g., MIMIC datasets \cite{johnson2016mimic, johnson2023mimic}) has provided valuable real-world data, allowing researchers to develop more advanced and complex deep learning models~\cite{LuRCKN21, lu2022context, poulain2024graph, JiangXC024, hu2024dualmar} to uncover predictive patterns from a data science perspective. These models often integrate domain knowledge of medical concepts to identify intricate correlations in disease progression and comorbidities, demonstrating promising predictive performance.
Despite their success in prediction accuracy, these methods have two key limitations: \begin{itemize}[leftmargin=1em, itemsep=0pt, topsep=0pt, parsep=0pt, partopsep=0pt]
    \item \textbf{Lack of interpretability:} deep learning models inherently function as ``black boxes'', offering little transparency into the clinical reasoning behind their predictions. 
    \item \textbf{Lack of interactivity:} most models operate in an end-to-end manner, limiting practitioner interaction with the system. This prevents users from asking follow-up questions, customizing prediction goals, or incorporating their own knowledge and experience to refine predictions.  
\end{itemize}

The lack of both \textbf{interpretability} and \textbf{interactivity} undermines trust and acceptance among healthcare professionals who depend on these predictions for informed decision-making.
In recent years, Large Language Models (LLMs) have demonstrated extensive knowledge, strong instruction-following capabilities, and impressive reasoning abilities, offering promising solutions to address these limitations. The development of agentic LLMs~\cite{wang2024colacare, Kim20224-MDAgents, zuo2024kg4diagnosis} has further enhanced their flexibility and capabilities through dynamic interactions with the environment, tool utilization, and inter-agent collaboration. These advancements hold great potential for enabling clinicians to engage more effectively with predictive models, fostering greater adaptability and user-driven refinement.

Inspired by these advancements, we propose \model{}, a knowledge-enhanced \textbf{A}gentic \textbf{C}ausal Discovery framework designed for \textbf{I}nterpretable and \textbf{I}nteractive \textbf{D}iagnosis Prediction. \model{} is a multi-agent system comprising three LLM agents namely Knowledge Synthesis Agent, Casual Discovery Agent, and Decision-Making Agent collaboratively, and is powered by both clinical dataset and domain knowledge. 
Similar to other deep learning approaches, \model{} predicts medical diagnoses by addressing the question:\textit{``What diseases is a patient likely to be diagnosed with given their past diagnosis history?"} However, unlike purely data-driven methods, \model{} approaches the problem from a causal perspective, delving deeper into the underlying mechanisms to answer:\textit{``What diseases are likely to be caused by the conditions a patient has already been diagnosed with?"}—thus reframing the task as a causal discovery problem. Recent advances in Large Language Models (LLMs) have demonstrated promising performance in causal discovery, alleviating the need for complex, data-centric, and resource-intensive traditional methods. However, LLMs often generate incorrect answers when domain knowledge is insufficient. To address this limitation, we enhance the causal discovery process by integrating both knowledge-driven reasoning through Retrieval Augmented Generation (RAG) and data-grounded inference, ensuring a deeper contextual understanding and better alignment with real-world observations.

To this end, we emphasize that \model{} is clinician-friendly framework that ensures both interpretability and interactivity.

\begin{itemize}[leftmargin=1em, itemsep=0pt, topsep=0pt, parsep=0pt, partopsep=0pt]
    \item \textbf{\model{} is interpretable.} The LLM agents
make \model{} inherently interpretable by enabling the decision-making agent to provide detailed explanations and reasoning behind its predictions. Additionally, \model{} gains an extra layer of interpretability through causal analysis. The causal graph generated
by the causal discovery agent offers an intuitive and comprehensive representation of the causal mechanisms between diseases, making it easier for users to
understand the underlying relationships.

\item \textbf{\model{} is interactive.} Clinicians can interact with and customize the prediction process through two pathways. First, the RAG feature of \model{} enables generalization by incorporating external knowledge, allowing clinicians to conveniently provide the knowledge sources their own or selected knowledge sources as the knowledge database. Second, clinicians can interact with the decision-making agent by specifying their personal preferences, ensuring that predictions are tailored to their specific needs.
\end{itemize}

We evaluate \model{} on EHR datasets, including MIMIC-III and MIMIC-IV, demonstrating superior performance along with enhanced interpretability and interactivity, supported by extensive ablation and case studies.

\begin{figure*}[h]
  \centering
  \includegraphics[width=\linewidth]{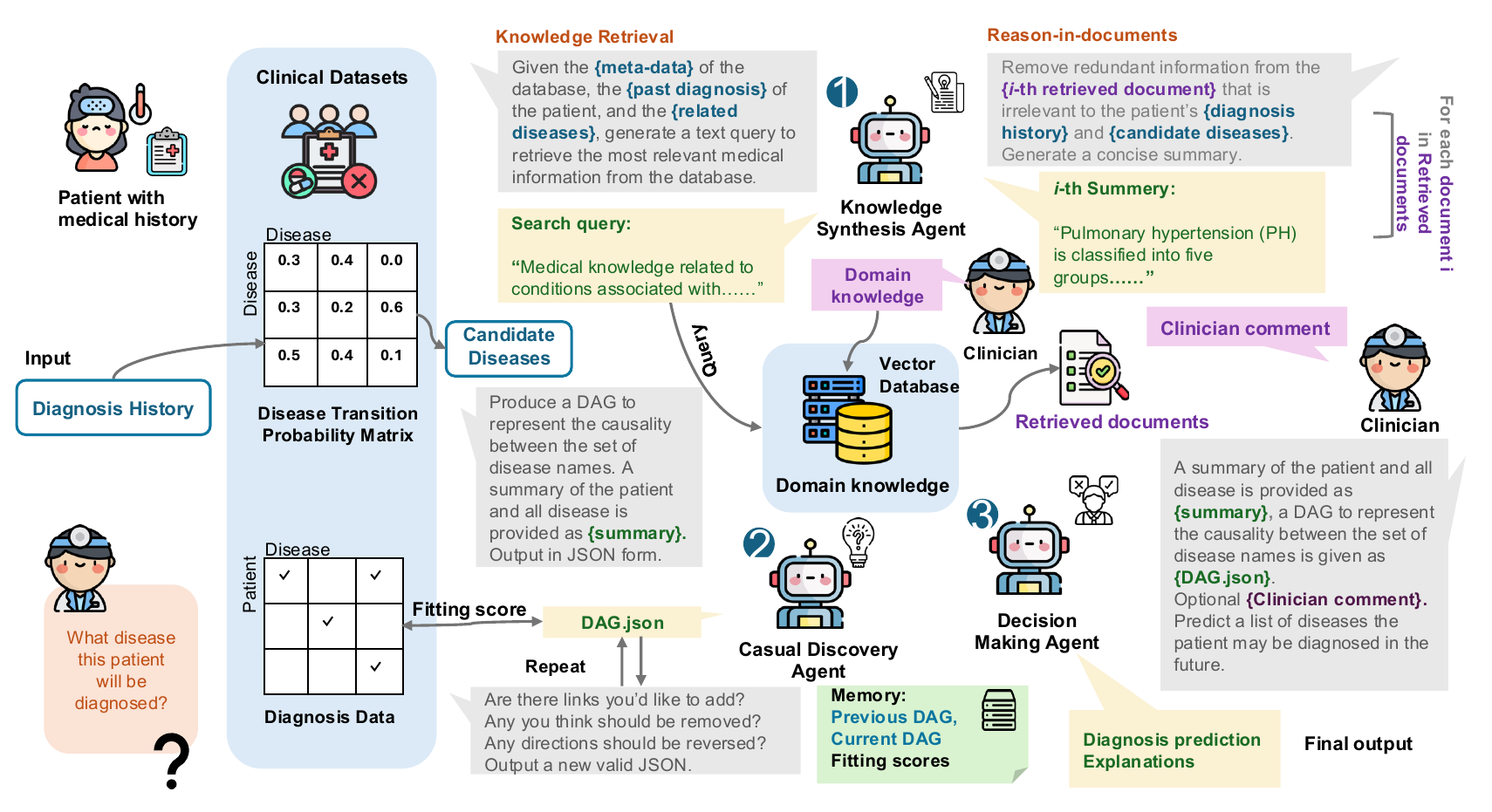}
  \caption{An overview of \model{} framework. It consists of three LLM-based agents working collaboratively and is powered by both clinical datasets and domain knowledge. During inference, a patient's diagnosis history is processed to identify possible diseases. A Knowledge Synthesis Agent retrieves and summarizes relevant documents. Then, a Causal Discovery Agent uncovers causal relationships using both external knowledge and observational data, forming a causal graph. Finally, a Decision-Making Agent integrates all this information—along with optional clinician input—to predict the diagnosis and provide explanations.}
  \label{fig:overview}
\end{figure*}

\section{Methodology}
We propose, \model{}, a multi-agent system consisting of three LLM-based agents working collaboratively and is powered by both clinical datasets and domain knowledge. \model{} aims to uncover causal relationships between diseases and predict future medical diagnoses. In this section, we introduce each LLM agent and knowledge module and provide a summary of the overall framework.

\subsection{Knowledge databases}
\subsubsection{Clinical datasets}
\label{section:data}
We construct a clinical dataset using training data from Electronic Health Records (EHR). Each record contains diagnosis information for individual patients across multiple visits. This database comprises two data frames: a Disease Transition Probability Matrix and a Diagnosis Matrix.

Let \(\mathcal{D}\) denote the complete set of diseases, and \(\mathcal{P}_{tr}\) denote the patient set from training data.
The Disease Transition Probability Matrix, denoted as \(\mathbf{A}_{T} \in \mathbb{R}^{\lvert \mathcal{D} \rvert \times \lvert \mathcal{D} \rvert}\), captures the probability of disease \( j \) occurring after disease \( i \). The underlying intuition is that temporal precedence is a necessary but not sufficient condition for causality. Identifying diseases that frequently follow a target disease helps narrow down potential causal candidates. By pre-selecting frequently co-occurring diseases, we provide the LLM agent with a shortlist of candidates, reducing its workload when assessing causal relationships, as we will discuss later. In constructing this matrix, for each visit, we define disease B as a successor of disease A if:
\begin{itemize}[leftmargin=1em, itemsep=0pt, topsep=0pt, parsep=0pt, partopsep=0pt]
    \item Disease B appears in a patient’s next visit after the visit in which disease A is diagnosed.
    \item Disease B appears in the same visit as disease A. 
\end{itemize}
   The second condition accounts for the fact that patients may not visit clinicians frequently, meaning that a succession disease and the target disease could be diagnosed simultaneously.

\begin{equation}
\label{a_t}
\mathbf{A}_{T}[a,b] = \frac{\mathcal{N}_{a,b}}{
\sum_{p \in \mathcal{P}} \sum_{i=1}^{m_{p-1}} \mathbb{I}[a \in \mathcal{D}_p^i]},
\end{equation}
where \begin{equation}
\mathcal{N}_{a,b} = \sum_{p \in \mathcal{P}_{\text{tr}}} \sum_{i=1}^{m_{p-1}} \mathbb{I}[a \in \mathcal{D}_p^i \land (b \in \mathcal{D}_p^{i+1} \lor b \in \mathcal{D}_p^i)],\end{equation}
and 
\(\mathbb{I}[\cdot]\) is the indicator function.
\(\mathbf{A}_{T}[a,b]\) is an entry of \(\mathbf{A}_{T}\), representing the transition probability between disease \(a\) and disease \(b\). \(\mathcal{P}_{tr}\) denotes the set of all patients in the training set, and \(\mathcal{D}_p^{i}\) represents the set of diseases diagnosed for patient \(p\) during their \(i\)-th visit. Note that \(\mathbf{A}_{T}\) is not necessarily symmetric, meaning that \(\mathbf{A}_{T}[a,b] \neq \mathbf{A}_{T}[b,a]\) in general.
The diagnosis matrix $\mathbf{A}_{D}\in\mathcal{R}^{\lvert \mathcal{P}_{tr} \rvert \times \lvert \mathcal{D} \rvert}$ records the occurrence of each disease for all patients. We consider the occurrence of disease $a$ for a patient to be 1 if the patient is has been diagnosed with the disease in any revisits:
\begin{equation}
\label{a_d}
  \mathbf{A}_{D}[p, a] = \mathbbm{1} (a\in \bigcup_{i \in m_p} \mathcal{D}_p^i ),
\end{equation}
we calculate the fitting score between the diagnosis matrix and the output causal graph to provide feedback to the causal discovery agent, as we will discuss in section \ref{a_causal}.

\subsubsection{Domain knowledge database}
\label{section:knowledge}
We construct a vector database powered by  ChromaDB\footnote{https://pypi.org/project/chromadb/}
as the source of external knowledge for the Retrieval-Augmented Generation (RAG) of the knowledge synthesis agent, as discussed in Section \ref{a_knowledge}.
The database can contain any domain knowledge from different sources such as web pages, published papers, or clinical notes. 
In this paper, we scrape text from Wikipedia pages corresponding to each disease listed in ICD-9. Each Wikipedia page is segmented into sections such as ``\texttt{Overview}'', ``\texttt{Signs and Symptoms}'', ``\texttt{Causes}'', ``\texttt{Diagnosis}'', ``\texttt{Prevention}'', ``\texttt{Treatment}'', ``\texttt{Epidemiology}'', ``\texttt{History}'', ``\texttt{Terminology}'', and ``\texttt{Society and Culture}''. When creating the vector database, each section is treated as an individual document, and its key vector is generated by embedding the document text using the pre-trained Sentence-BERT model \texttt{all-MPNet-base-v2}, which exhibits exceptional performance in capturing semantic similarities between sentences.

\subsection{Multi-agent Framework}

The goal of \model{} is to predict a patient's future diagnoses by conducting causal discovery on their diagnosis history and identifying diseases that are most likely caused by past conditions. However, directly asking an LLM agent to perform this task across thousands of diseases would be computationally expensive. Instead, we leverage a Disease Transition Probability Matrix, denoted as \(\mathbf{A}_{T}\), to select candidate diseases, acknowledging that temporal precedence is a necessary condition for causality.  For a patient \( p \), let \( D^p \) denote the set of diseases they have been diagnosed with in the past. The set of candidate diseases \( S^p \) that could be caused by \( D^p \) is then obtained as:  
\begin{equation}
\label{s_p}
S^p = \{ b \mid \mathcal{M}[a, b] > \epsilon, \quad \forall a \in D_p \}
\end{equation}

We then provide both the diagnosis history set \( D_p \) and the candidate disease set \( C_p \) to the agents to determine which diseases are causally linked. To ensure that the causal discovery process is grounded in sufficient domain knowledge, a straightforward approach would be to query a vector database separately for each disease in \( D_p \) and \( C_p \) and send the retrieved text to the causal discovery agent.  However, this approach has two major drawbacks:  1) Independently querying each disease focuses on individual diseases rather than the relationships between them, failing to retrieve information most relevant to causal links.  
2) The retrieved documents may contain redundant information, be excessively long, and exceed the processing capacity of LLMs.  
To address these issues, we develop a Knowledge Synthesis Agent.

\textbf{Knowledge Synthesis Agent, \( \mathcal{A}_{\text{knowledge}} \).}  
\label{a_knowledge}
The role of \( \mathcal{A}_{\text{knowledge}} \) is to generate high-quality contextual information for the causal discovery process. Its generation process consists of two steps.  
In the first step, the agent is provided with the database metadata, the patient's diagnosis history \( D_p \), and the candidate disease set \( C_p \). It is responsible for generating a query text to retrieve relevant information from the database. This query text should effectively summarize \( D_p \) and \( C_p \) while being tailored to the specific database based on its metadata, which defines its characteristics and content. We then encode the query text using the same pre-trained Sentence-BERT model and retrieve the \( k \) most relevant documents.  
In the second step, \( A_{ks} \) performs \textit{reasoning-in-documents}, refining the retrieved information by removing redundancies and generating a concise summary. These summarized documents are then stored for use by the causal discovery agent, enabling a \textit{Retrieval-Augmented Generalization (RAG)} approach. We summarize the workflow of the \( \mathcal{A}_{\text{knowledge}} \) in Algorithm \ref{algo:knowledge}. 

\textbf{Causal Discovery Agent, \( \mathcal{A}_{\text{causal}} \).}  
\label{a_causal}
The role of \( \mathcal{A}_{\text{causal}} \) is to identify potential causal relationships among a set of diseases. We provide it with the patient's diagnosis history set \( D_p \) and the candidate disease set \( C_p \) as a whole, along with the summarized external knowledge generated by \( \mathcal{A}_{\text{knowledge}} \).

We then adapt the iterative causal discovery procedure proposed in \cite{abdulaal2024causal}:  
\begin{enumerate} [leftmargin=1em, itemsep=0pt, topsep=0pt, parsep=0pt, partopsep=0pt]
    \item \textit{Hypothesis generation.} Given the summarized external knowledge and the empty graph \( \mathcal{G}^{\emptyset} \), which consists of all entities in \( D_p \) and \( C_p \) with no initial relations, the \( \mathcal{A}_{\text{causal}} \) LLM generates an initial causal graph as a directed acyclic graph (DAG), \( \mathcal{G}^{s}_{t=0} \).  
    \item \textit{Model fitting.} At each iteration \( t \), we fit the causal model using a data-driven approach with real-world observations. Specifically, we measure the fitting score as the log-likelihood, \( l_t \), of the diagnosis matrix \( \mathbf{A}_{D} \) under the current model \( \mathcal{G}^{s}_{t} \).  

   \begin{equation}
    \begin{split}
l_{t} = \sum_{p\in\mathcal{P}_{tr}} \sum_{a\in\mathcal{D}} 
    \log P(X_p^{a} \mid \{X_p^{b} \mid b \in \mathbf{Pa}(a)\})
    \end{split}
    \end{equation}

where \(\mathbf{Pa}(a)\) denotes the 
parent diseases of \(a\) in \(\mathcal{G}^{s}_{t}\), and \(X_p^{a} \in \{0,1\}\) represents the observation of disease \(a\) in patient \(p\).
    
    \item \textit{Post-processing.} We update the memory \( M_t \) to store the causal graph and the fitting score from the previous and current time steps, including \( M_t \), \( M_{t-1} \), \( l_t \), and \( l_{t-1} \). This memory is retained for the next step.  
    \item \textit{Hypothesis amendment.} The LLM refines the causal model based on the stored memory to enhance its accuracy and better capture causal relationships. It then outputs the updated causal graph as \( \mathcal{G}^{s}_{t+1} \).  
\end{enumerate}

Steps 2 to 4 are repeated iteratively until a stopping criterion is met, (when the change in \( \mathcal{G}^{s}_{t} \) falls below a predefined threshold or the number of iterations exceeds a limit). We summarize the workflow of the Causal Discovery Agent in Algorithm \ref{algo:causal}.

\textbf{Decision-Making Agent, \( \mathcal{A}_{\text{dm}} \).}
 The role of \( \mathcal{A}_{\text{dm}} \) is to integrate and evaluate all available information, including diagnosis history sets, summarized knowledge, and the causal graph, to make the final prediction on a patient's diagnosis. Additionally, clinicians or users can provide their preferences, comments, or experiences to customize the prediction. For example, they may indicate that they are particularly concerned about kidney-related diseases. The agent is then tasked with producing the diagnosis list in a structured format along with an explanation of the reasoning behind its decision. We summarize the workflow of the Decision-Making Agent in Algorithm \ref{algo:decision}.

\subsection{\model{} Inference}

\model{} does not involve any training process but requires data preprocessing. First, the EHR training dataset is processed to construct the Disease Transition Probability Matrix \(\mathbf{A}_{T}\) and the Diagnosis Matrix \(\mathbf{A}_{D}\), as described in Section \ref{section:data}. Additionally, a knowledge vector database \(\Gamma\) is prepared following Section \ref{section:knowledge}. Both matrices and the database are stored for later inference.  During inference, for each patient, we collect their diagnosis history \(\mathcal{D}_p\) and apply the preprocessing steps outlined in Section \ref{section:data} to determine the candidate disease set \(\mathcal{S}_p\). The Knowledge Synthesis Agent \(\mathcal{A}_{\text{knowledge}}\) then retrieves relevant documents from \(\Gamma\) and summarizes them into \(\Gamma_p^{\text{summary}}\).  Next, the Causal Discovery Agent \(\mathcal{A}_{\text{causal}}\) iteratively uncovers causal relationships within the expanded disease set \(\mathcal{D}_{p} \cup \mathcal{S}_{p}\), leveraging both external knowledge from \(\Gamma_p^{summary}\) and observational data from \(\mathbf{A}_{D}\). This process results in a causal graph \(\mathcal{G}^s\).  Finally, the Decision-Making Agent \(\mathcal{A}_{\text{dm}}\) integrates all available information, including the diagnosis history \(\mathcal{D}_p\), candidate diseases \(\mathcal{S}_p\), summarized documents \(\Gamma_p^{summary}\), causal graph \(\mathcal{G}^s\), and an optional clinician-provided comment \(\mathcal{C}\). Using this information, the model predicts the patient's diagnosis and provides explanations for the decision.  We provide the overview of \model{} in Figure \ref{fig:overview}, the prompt details in Appendix \ref{apd:prompt}. 
The workflow of its inference process is shown in Algorithm \ref{algo:acid}.

\section{Experiments \& Setup}

\subsection{Datasets}
We utilize both the MIMIC-III~\citep{johnson2016mimic} and MIMIC-IV~\citep{johnson2023mimic} datasets for our experiments. 
MIMIC-III contains 7,493 patients with multiple visits ($T \geq 2$) between 2001 and 2012, while MIMIC-IV includes 85,155 patients with multiple visits spanning from 2008 to 2019.
Due to the overlapping time period between the two datasets, we randomly sample 10,000 patients from MIMIC-IV between 2013 and 2019 to ensure minimal redundancy.
For the diagnosis prediction task, the objective is to predict the medical codes appearing in the subsequent admission.

To verify the efficiency of the proposed model, MIMIC-III is split into training (6,000 patients), validation (1,900 patients), and test (1,000 patients) sets. 
Similarly, MIMIC-IV is divided into 8,000, 1,000, and 1,000 patients accordingly. 
The last recorded visit of each patient serves as the prediction target, while the preceding visits are used as input features.
Different from typical predictive models, we feed \model{} by admission records of those patients in the training data for getting co-occurrence matrix, and examine predictive performance upon 500 patient cohort.

\subsection{Tasks \& Evaluation Metrics}
Our experiments focus on the task of \textit{Diagnosis Prediction}, which aims to predict all medical codes that will appear in a patient's next admission. This task is formulated as a multilabel classification problem.
To evaluate model performance, we use weighted F$_1$ score (w-F$_1$) and top-$k$ recall (R@$k$) as metrics, following prior work~\citep{choi2016doctor, bai2018interpretable}. The w-$F_1$ score is a weighted sum of the $F_1$ score across all classes, providing an overall assessment of prediction quality. The R@$k$ metric represents the proportion of true-positive instances among the top-$k$ predictions relative to the total number of positive samples, reflecting model effectiveness in capturing relevant medical codes.

\subsection{Baselines}
To assess the performance of \model{}, we compare it against 8 machine learning (ML)-based  EHR models originally designed for diagnostic prediction: 
(i) RNN/CNN-based models: RETAIN~\citep{choi2016retain}, Dipole~\citep{ma2017dipole}, and Timeline~\citep{bai2018interpretable}. 
(ii) Graph-based models: Chet~\citep{lu2022context} and SeqCare~\citep{xu2023seqcare}. 
(iii) Transformer-based models: G-BERT~\citep{ijcai2019p825}, BEHRT~\citep{li2020behrt}, and GT-BEHRT~\citep{poulain2024graph}.

Moreover, we compare 3 most recent baselines that combine language models with machine learning-based predictors (LM+ML): GraphCare~\cite{JiangXC024}, RAM-EHR~\cite{xu2024ram}, and DualMAR~\citep{hu2024dualmar}.
To ensure a fair comparison, we use condition codes as the sole input feature (e.g., excluding procedures and medications used in GraphCare), and we reconstruct the knowledge base using ICD-9-CM codes instead of CCS codes.
Other agentic baselines like ColaCare~\cite{wang2024colacare} are excluded from the comparison due to its extensive input requirements, which lead to unstable predictions when only condition codes are provided.

\begin{table*}[t]
    \small
    \centering
    \caption{\textbf{Prediction Results on MIMIC-III and MIMIC-IV for Diagnosis Prediction.} 
    The best results for each metric are \textbf{bold}, and the second bests results are highlighted in gray.}
    \label{tab:mimic_main}
    \begin{tabular}{clccccccc}
        \toprule
         & & \multicolumn{3}{c}{\textbf{MIMIC-III}}& & \multicolumn{3}{c}{\textbf{MIMIC-IV}} \\
        \textbf{Type}& \textbf{Models} & \textbf{w-}$\mathbf{F_1}$ & \textbf{R@10} & \textbf{R@20} & &\textbf{w-}$\mathbf{F_1}$ & \textbf{R@10} & \textbf{R@20} \\
        \midrule
        \multirow{8}{*}{ML} 
        & RETAIN & 18.37 (0.78) & 32.12 (0.79) & 32.54 (0.63) && 23.11 (0.78) & 37.32 (0.79) & 40.15 (0.63) \\
        & Dipole & 14.66 (0.21) & 28.73 (0.22) & 29.44 (0.21) & &22.16 (0.21) & 36.21 (0.22) & 38.74 (0.21) \\
        & Timeline & 20.46 (0.22) & 30.73 (0.12) & 34.83 (0.10) && 24.76 (0.22) & 39.89 (0.12) & 44.87 (0.10) \\
        & Chet & 22.63 (0.22) & 33.64 (0.32) & 37.87 (0.22) & &25.74 (0.22) & 39.23 (0.32) & 42.67 (0.22) \\
        & SeqCare & 24.36 (0.12) & 37.47 (0.11) & 40.53 (0.12) & &26.12 (0.12) & 42.91 (0.11) & 46.25 (0.12) \\
        & G-BERT & 22.28 (0.32) & 35.62 (0.21) & 36.46 (0.22) & &25.12 (0.32) & 41.91 (0.21) & 46.25 (0.22) \\
        & BEHRT & 23.15 (0.21) & 34.68 (0.32) & 35.97 (0.11) && 24.53 (0.21) & 38.42 (0.32) & 44.89 (0.11) \\
        & GT-BEHRT & 25.21 (0.18) & 36.15 (0.23) & 40.97 (0.41) && \textbf{30.17 (0.18)} & \cellcolor{gray!18}44.93 (0.23) & \cellcolor{gray!18}50.67 (0.41) \\
        \cmidrule{1-9}
        \multirow{3}{*}{LM+ML}
        & GraphCare & 25.16 (0.31) & 36.74 (0.28) & \cellcolor{gray!18}41.89 (0.36) & &27.59 (0.31) & 42.07 (0.28) & 48.19 (0.36) \\
        & RAM-EHR & 23.27 (0.24) & 34.66 (0.18) & 38.49 (0.25) & &26.97 (0.29) & 41.17 (0.30) & 46.23 (0.21) \\
        & DualMAR & \cellcolor{gray!18}25.37 (0.17) & \cellcolor{gray!18}38.24 (0.26) & 41.86 (0.24) & &27.97 (0.17) & 44.07 (0.26) & 48.19 (0.24) \\
        \cmidrule{1-9}
        \multirow{1}{*}{Agent}
        & \model{} & \textbf{28.61 (0.00)} & \textbf{38.52 (0.00)} & \textbf{43.86 (0.00)} &&\cellcolor{gray!18}29.87 (0.00) & \textbf{45.66 (0.00)} & \textbf{51.73 (0.00)} \\
        \bottomrule
    \end{tabular}
\end{table*}

\subsection{Implementation Details}
We implement \model{} using \texttt{Python 3.10}. For all agents, we utilize \texttt{ChatGPT-4o mini}~\cite{openai2024gpt4technicalreport}, accessed via the \texttt{Azure OpenAI}, as our LLM. To build the vector database, we employ \texttt{ChromaDB}, where document embeddings are generated using a pre-trained Sentence-BERT \texttt{all-MPNet-base-v2} model provided by the \texttt{Sentence Transformers}. 
We report the average performance (\texttt{\%}) and standard deviation of each baseline over 5 runs, and we set the temperature value in \model{} as 0.
When evaluating the prediction performance of \model{} we set the optional clinical comment to be empty. 

\subsection{Main Results}
Table~\ref{tab:mimic_main} presents the performance comparison, demonstrating that the proposed model, \model{}, achieves state-of-the-art results across both datasets. Specifically, \model{} outperforms GT-BEHRT by 2.44\% in w-F$_1$, 0.73\% in R@10, and 1.06\% in R@20 on MIMIC-IV, with similar performance gains observed on MIMIC-III.
The results further indicate that graph-based and transformer-based models consistently outperform RNN- and CNN-based approaches. 
Notably, knowledge-based models such as DualMAR leverage knowledge graphs to enhance learning, yielding a 9\% improvement in R@20 on MIMIC-III. Similarly, transformer-based models like GT-BEHRT improve w-F$_1$ by approximately 8\% on MIMIC-IV. While GT-BEHRT and DualMAR achieve competitive performance in certain metrics, \model{} consistently surpasses both across the majority of evaluation criteria.
Overall, these findings underscore the effectiveness of \model{} in diagnosis prediction and highlight the potential of a unified agentic framework for advancing predictive healthcare.

\begin{figure}[t]
  \centering
  \includegraphics[width=\linewidth]{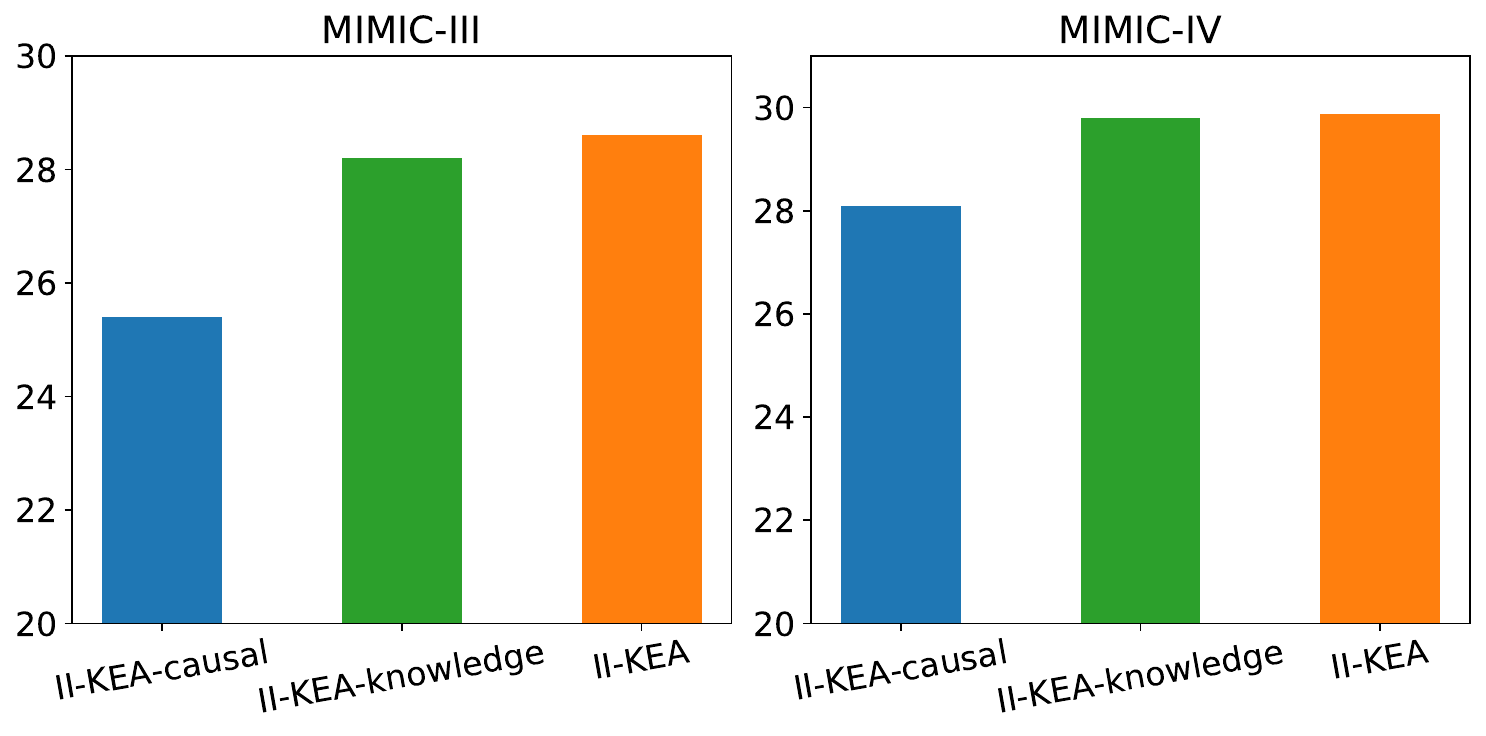}
  \caption{Comparison between different version of \model{}. F1 scores on MIMIC-III and MIMIC-IV are reported.}
  \label{fig:ablation}
\end{figure}

\begin{figure*}[h]
  \centering
  \includegraphics[width=\linewidth]{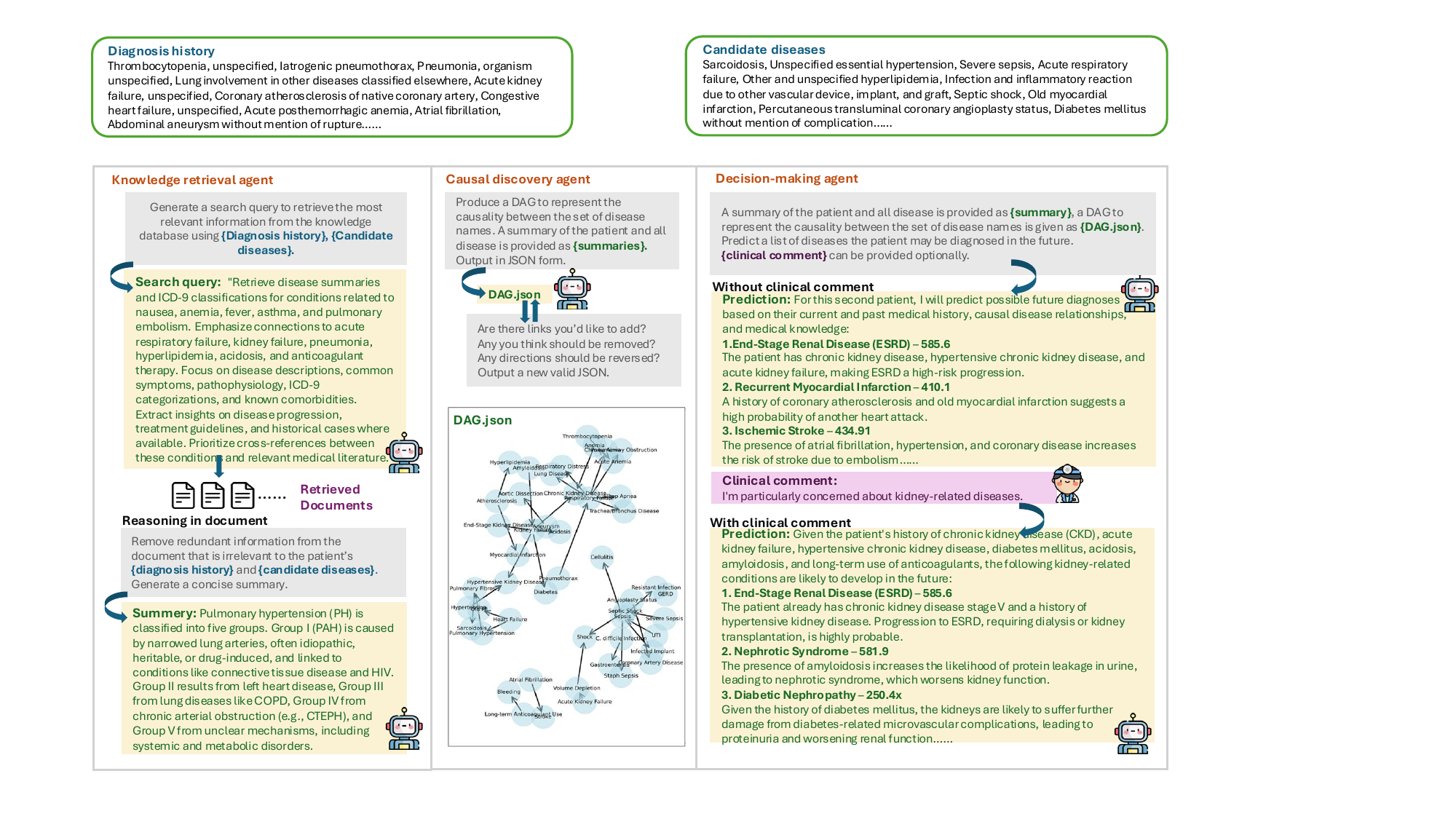}
  \caption{A case study on a patient from MIMIC-III.}
  \label{fig:case-study}
\end{figure*}

\subsection{Ablation Study}
We conduct ablation studies to evaluate the effectiveness of components in \model{}. 
Specifically, we aim to understand how causal analysis and external knowledge contribute to prediction performance. 
The Knowledge Synthesis Agent and the Causal Discovery Agent are separately removed from the prediction workflow and weighted F1 scores are reported in Figure \ref{fig:ablation}.
We denote the version without the Causal Discovery Agent as \model{}-\textit{causal} and the version without the Knowledge Synthesis Agent as \model{}-\textbf{knowledge}. 
The results show that \model{}-\textit{causal} experiences a more significant performance drop than the full model, highlighting the crucial role of causal analysis in improving prediction accuracy. 
In contrast, removing external knowledge (\model{}-\textbf{knowledge}) results in only a marginal decline, suggesting a lesser impact in its current form.
We hypothesize that this is because our knowledge database is sourced from Wikipedia, which primarily serves as a demonstration of external knowledge integration but may offer limited domain-specific medical insights. 
However, this also underscores the potential for improvement by incorporating curated databases or clinician-maintained knowledge sources.

\subsection{Case Study}

We conduct a case study to analyze how different agents within \model{} function and collaborate during the decision-making process. We randomly select a patient from the MIMIC-III dataset and report the output of each agent during inference, as shown in Figure \ref{fig:case-study}. The Causal Discovery Agent identifies a causal graph, visualized in the figure, which helps illustrate the underlying mechanisms connecting different diseases.
For the Decision-Making Agent, we provide outputs both with and without clinician input. In the first query, no specific guidance is given, leading to a more general prediction that considers all possible diseases. In contrast, in the second query, the clinician provides additional input, specifying a focus on kidney-related diseases. Consequently, the model prioritizes kidney-related predictions.
It is important to note that the performance of these two versions cannot be directly compared; rather, the key advantage is that clinicians can incorporate their expertise and preferences to tailor predictions to their specific needs (e.g., a nephrologist may prioritize kidney-related diseases). 
We also observe that both versions of the predictions not only provide disease codes but also offer detailed explanations, enhancing interpretability and helping clinicians in making informed decisions and determining next-step treatment plans.

\section{Related Work}
\label{sec:floats}

We categorize prior work into clinical prediction (section~\ref{sec:related_health}), agentic approaches (section~\ref{sec:related_agent}), and causal inference (section~\ref{app:causal_related}).

\subsection{Predictive Healthcare in EHR}
\label{sec:related_health}

Predictive modeling in healthcare has advanced significantly with the adoption of deep learning techniques~\citep{badawy2023healthcare, 10.1145/3584371.3613008} applied to EHR data. Existing neural network-based models, including RNN/Attention-based approaches~\citep{choi2016doctor, choi2017gram, ma2020adacare}, graph-based models~\citep{choi2017gram, ma2018kame, LuRCKN21}, and Transformer-based architectures~\citep{ShangMXS19, luo2020hitanet, poulain2024graph}, have demonstrated effectiveness in capturing temporal patterns and interactions among medical concepts. Recent work~\citep{JiangXC024} has explored leveraging external knowledge sources beyond hierarchical structures such as ICD-9-CM by integrating Large Language Models (LLMs) to enhance medical predictions.

Still, most models remain black boxes~\citep{ zhang2024natural}, offering limited interpretability and restricting healthcare professionals from interacting with the system to refine or adjust predictions. In clinical applications, predictive models must provide faithful explanations, such as causal pathways, and allow interactive refinement based on expert guidance.

\subsection{LLM Agents for Healthcare AI}
\label{sec:related_agent}

More recently, LLMs have demonstrated agentic capabilities in clinical applications through multi-agent frameworks. 
EHRAgent~\citep{shi2024ehragent} utilizes multiple agents for multi-tabular retrieval, integrating external tools and long-term memory to handle complex clinical queries. 
KG4Diagnosis~\citep{zuo2024kg4diagnosis} enhances diagnostic reasoning through hierarchical agent collaboration and knowledge graph construction guided by semantic understanding. ColaCare~\citep{wang2024colacare} improves EHR-based report generation and treatment planning by facilitating collaboration between DoctorAgents and MetaAgent using retrieval-augmented generation (RAG) techniques. 
MDAgents~\citep{Kim20224-MDAgents} and AgentClinic~\citep{agentclinicbenchmark} simulate clinical interactions using multi-agent systems, where agents collaborate to support multi-modal reasoning and communication benchmarking.
These studies highlight an emerging trend in agentic AI for clinical applications, where LLMs leverage in-context learning and dynamically retrieve medical knowledge to provide personalized and adaptive responses.

Still, the integration of LLM agents for sequential diagnostic prediction remains underexplored, presenting an opportunity to develop interactive and explainable models for medical diagnosis.

\subsection{Causality Inference on LLM}
\label{app:causal_related}

Causal inference is a cornerstone of medical research, enabling the discovery of relationships between clinical factors. LLMs, equipped with extensive domain knowledge, have the potential to assist in causal graph generation and infer causal relationships from unstructured data. Consequently, recent studies have started exploring LLM-driven causal discovery frameworks~\citep{liu2024large}. 
Several works~\citep{le2024multi, shen2024exploring, choi2022lmpriorspretrainedlanguagemodels, kıcıman2024causalreasoninglargelanguage, long2023causaldiscovery, jiralerspong2024efficientcausalgraphdiscovery} have employed LLMs for causal relation inference and graph generation, yet their application to EHR-based predictive tasks remains limited. Most existing approaches focus on general causal reasoning tasks or static datasets, without fully leveraging the interactive and adaptive capabilities of LLM agents for healthcare-specific causal discovery.

However, causal discovery is essential for diagnosis prediction, as it provides a structured explanation of disease co-occurrences, facilitating more transparent and interpretable decision-making. Bridging this gap is crucial for achieving explainable AI in clinical practice by enabling collaborative causal reasoning among AI agents. Furthermore, integrating interactive causal discovery mechanisms allows healthcare professionals to refine insights and better understand disease relationships, ultimately improving diagnosis and treatment planning.

\section{Computation Cost}
We evaluated the cost of performing one prediction loop with \model{} using a sample of 10 patients. On average, each patient required 7,803 tokens (4,871 input tokens and 2,932 output tokens), resulting in an estimated cost of approximately 0.00263 USD per patient when using GPT-4o Mini (at rates of 0.0003 USD per 1,000 input tokens and 0.0004 USD per 1,000 output tokens). The average processing time was 87 seconds per patient.

\section{Privacy and Ethical Statement}
\label{app:ethical}

Our work involves the analysis of EHR data, which contains sensitive personal medical information. 
In compliance with the PhysioNet Credentialed Health Data Use Agreement 1.5.0\footnote{https://physionet.org/about/licenses/physionet-credentialed-health-data-license-150/}, we conducted all interactions between the language models and the EHR data through \textit{Azure OpenAI Service}\footnote{https://azure.microsoft.com/en-us/products/ai-services/openai-service/}, which adheres to enterprise-grade security and compliance standards. 
We also submitted the opt-out form\footnote{https://azure.microsoft.com/en-us/products/cognitive-services/openai-service/} to decline human review in terms of the responsible use guidelines specified for MIMIC datasets available at \textit{Responsible Use of MIMIC Data with online services}\footnote{https://physionet.org/news/post/gpt-responsible-use}, which outlines proper handling of EHR data when used with generative models. 
This ensured that the capabilities of large language models were applied without compromising the privacy and confidentiality of patient information.
Furthermore, we continuously and carefully monitor our compliance with these guidelines and relevant privacy regulations to uphold the ethical use of data in our research and operations.

\section{Conclusion}
In this paper, we introduce \model{}, a knowledge-enhanced Agentic Causal Discovery framework designed for interpretable and interactive diagnosis prediction. \model{} consists of three LLM-based agents working collaboratively and is powered by both clinical datasets and domain knowledge. We evaluate \model{} on the MIMIC-III and MIMIC-IV datasets and conduct both ablation and case studies to demonstrate its effectiveness.
The ethical consideration can be checked in Appendix~\ref{app:ethical}.

\section{Limitation and Future Work}
\model{} showcases a promising paradigm for interpretable and interactive diagnosis prediction by leveraging LLM agents.
While \model{} focuses on diagnosis prediction, future research will explore to solve a broader range of medical challenges tailored to clinicians' needs, such as treatment planning and personalized medical recommendations.
Future work includes:
\begin{itemize}[leftmargin=1em, itemsep=0pt, topsep=0pt, parsep=0pt, partopsep=0pt]
    \item \textbf{Enhancing external domain knowledge:} In this work, we use the Wikipedia database as a proof of concept. In future work, we aim to integrate more domain-specific external knowledge sources to enhance diagnosis prediction in fine-grained target domains.
    \item \textbf{Expanding task diversity:} While this work focuses on diagnosis prediction, future research will explore additional tasks tailored to clinicians' needs, such as treatment planning and personalized medical recommendations.
    \item \textbf{Incorporating multiple stakeholders:} The current version of \model{} facilitates interactions only with clinicians. Future iterations will explore collaborative decision-making involving multiple stakeholders to enhance holistic and patient-centered care.
\end{itemize}


\section*{Acknowledgment}

We thank the anonymous reviewers and area chairs for their valuable feedback. We used ChatGPT to polish our writing. This work was supported in part by the U.S. National Science Foundation under grants 2047843 and 2437621. Any opinions, findings, conclusions, or recommendations expressed in this material are those of the authors and do not necessarily reflect the views of the National Science Foundation.

\renewcommand{\thesection}{\Alph{section}}
\bibliography{main}
\setcounter{section}{0}
\newpage
\section*{Appendix}

\renewcommand{\thesection}{\Alph{section}}

\section{Potential Risk of AI clinical prediction}

There are three kinds of potential risks by using agentic AI for predictive healthcare:

\textbf{(1) Clinical Safety Risks.} AI clinical prediction systems pose significant clinical safety risks including diagnostic errors from overreliance on black-box models, perpetuation of healthcare biases present in training data, and poor generalization to populations underrepresented in datasets. These issues can lead to differential prediction accuracy across demographic groups and unreliable performance in real-world clinical scenarios.

\textbf{(2) Ethical \& Legal Risks.} Ethical and legal concerns arise from the lack of interpretability in traditional AI systems, creating accountability challenges when predictions contribute to adverse outcomes. Additionally, opaque AI decision-making may compromise informed consent and patient autonomy, while large-scale EHR data requirements raise privacy and security concerns regarding unauthorized access to sensitive medical information.

\textbf{(3) Operational Risks.} Operational challenges include disruption of established clinical workflows, potential skill atrophy among healthcare professionals who become overly dependent on AI recommendations, and false confidence generated by systems that provide predictions without uncertainty quantification. These issues particularly affect trainees and may lead to inadequate consideration of alternative diagnoses.

The II-KEA framework mitigates these risks through explicit causal reasoning and detailed explanations that enable clinician evaluation of AI recommendations, reducing overreliance and supporting accountability. Its interactive design requires active clinician participation, preserving clinical judgment while allowing healthcare professionals to inject domain knowledge and maintain diagnostic control. The multi-agent architecture provides validation layers, external knowledge integration reduces bias, and transparent reasoning supports both audit trails and informed consent processes.

\section{Pseudocodes}

The pseudocodes of three agents in \model{} are demonstrated in Algorithms~\ref{algo:knowledge} to~\ref{algo:decision}, and the Algorithm~\ref{algo:acid} shows the overall process of \model{}.

\begin{algorithm}
\caption{Knowledge Synthesis Agent, \( \mathcal{A}_{\text{knowledge}} \).}
\label{algo:knowledge}
\begin{algorithmic}[1]
\Statex \textbf{Input:} Diagnosis history $\mathcal{D}_p$, Candidate diseases $\mathcal{S}_p$, Knowledge vector database $\Gamma$ with meta data $\mathcal{K}_{\Gamma}$.
\Statex \textcolor{cyan}{// \textit{Generate search query}}
\Statex
\(
    q_{\text{search}} = LLM_{\text{search}}(\mathcal{D}_p, \mathcal{S}_p,\mathcal{K}_{\Gamma})
\)
\Statex \textcolor{cyan}{// \textit{ Retrieve $k$ most relevant documents from database $\Gamma$}}
\Statex
$\Gamma_{q} = query(q_{\text{search}}, \Gamma)$
\Statex {\textbf{for} each document $\gamma\in \Gamma^{q}$:} 
 \Statex  \hspace{1em} \textcolor{cyan}{// \textit{Reasoning in document}}
\Statex \hspace{1em} $a_\gamma = LLM_{\text{reason-in-doc}}(\gamma, \mathcal{D}_p, \mathcal{S}_p)$
\Statex \textbf{end for}
\Statex \textbf{Output:} Summarized document set $\Gamma_p^{\text{summary}} = a_\gamma \mid \gamma \in \Gamma^{q}$. 
\end{algorithmic}
\end{algorithm}

\begin{algorithm}
\caption{Causal Discovery Agent, \( \mathcal{A}_{\text{causal}} \).}
\label{algo:causal}
\begin{algorithmic}[2]
\Statex \textbf{Input:} Empty graph $\mathcal{G}^{\emptyset}$ consists of entities from  $\mathcal{D}_p \cup \mathcal{S}_p$, Candidate diseases ,  Summarized document set $\Gamma_p^{\text{summary}}$, diagnosis matrix $\mathbf{A}_{D}$

\Statex $t\triangleq0$
\Statex \textcolor{cyan}{// \textit{Hypothesis generation}}
\Statex
\(
\mathcal{G}^s_{t=0} = \text{LLM}_{\text{hypo-gen}} (\mathcal{G}^{\emptyset}, \Gamma_p^{\text{summary}})
\)

\Statex \textbf{While} 1: 

\Statex \hspace{1em} \textcolor{cyan}{// \textit{Model fitting}}
\Statex \hspace{1em} $l_t = $\texttt{log likelihood} $(\mathcal{G}^s_{t}, \mathbf{A}_{D})$
\Statex \hspace{1em} \textcolor{cyan}{// \textit{Post-processing}}
\Statex \hspace{1em} $\mathcal{M} \triangleq \{\mathcal{G}^s_{t}, \mathcal{G}^s_{t-1}, l_t, l_{t-1}\}$
\Statex \hspace{1em} \textcolor{cyan}{// \textit{Hypothesis amendment}}
\Statex \hspace{1em} \(
\mathcal{G}^s_{t+1} = \text{LLM}_{\text{hypo-amend}} (\mathcal{M})
\)
\Statex \hspace{1em} \textbf{if} \textit{stopping criteria} is meet: $\mathcal{G}^s \triangleq \mathcal{G}^s_{t+1}$
\Statex \hspace{2em} \textbf{break}
\Statex \hspace{1em} $t = t+1$
 \Statex \textbf{Output:} Final causal graph $\mathcal{G}^s$

\end{algorithmic}
\end{algorithm}

\begin{algorithm}
\caption{Decision-Making Agent, \( \mathcal{A}_{\text{dm}} \).}
\label{algo:decision}
\begin{algorithmic}[1]
\Statex \textbf{Input:} Diagnosis history $\mathcal{D}_p$, Candidate diseases $\mathcal{S}_p$, Summarized document set $\Gamma_p^{\text{summary}}$, Causal graph $\mathcal{G}^s$, Optional clinician comment $\mathcal{C}$
\Statex \textcolor{cyan}{// \textit{Make diagnosis prediction with explanations}}
\Statex $\mathcal{D}_{\text{pred}}, \mathcal{E} = \text{LLM}_{\text{decison}} (\mathcal{S}_p, \Gamma_p^{\text{summary}}, \mathcal{G}^s, \mathcal{C})$
\Statex \textbf{Output:} Predicted diagnosis $\mathcal{D}_{\text{pred}}$ and explanations $\mathcal{E}$
\end{algorithmic}
\end{algorithm}

\begin{algorithm}
\caption{\model{} inference.}
\label{algo:acid}
\begin{algorithmic}
\Statex \textbf{Require:} Pretrained LLM model, EHR training data $\mathcal{D}_p \mid p\in\mathcal{P}_{tr}$, Knowledge vector database $\Gamma$ with meta data $\mathcal{K}_{\Gamma}$. 

\Statex \textcolor{cyan}{// \textit{Data processing}}
\Statex Calculate $\mathbf{A}_{T}$ and $\mathbf{A}_{D}$ with Eqs. \ref{a_t} and \ref{a_d}. 

\Statex \textcolor{cyan}{// \textit{Inference}}

\Statex \textbf{For} $p \in \mathcal{P}_{test}$:
\Statex  \hspace{1em} \textbf{Input:} Diagnosis history $\mathcal{D}_p$ 
\Statex  \hspace{1em}  Obtain candidate disease $\mathcal{S}_p$ with Eq. \ref{s_p}.
\Statex  \hspace{1em} \textcolor{cyan}{// \textit{Knowledge Synthesis Agent}}
\Statex  \hspace{1em}  $\Gamma_p^{\text{summary}} = \mathcal{A}_{\text{knowledge}} (\mathcal{D}_p, \mathcal{S}_p,\mathcal{K}_{\Gamma})$
\Statex  \hspace{1em} \textcolor{cyan}{// \textit{ Causal Discovery Agent}}
\Statex  \hspace{1em}  $\mathcal{G}^s = \mathcal{A}_{\text{causal}} (\mathcal{D}_p, \mathcal{S}_p,\Gamma_p^{\text{summary}}, \mathbf{A}_{D})$
\Statex  \hspace{1em}  \textcolor{cyan}{// \textit{Decision-Making Agent}}
\Statex  \hspace{1em}  $\mathcal{D}_{\text{pred}}, \mathcal{E} = \mathcal{A}_{\text{dm}} (\mathcal{D}_p\mathcal{S}_p, \Gamma_p^{\text{summary}}, \mathcal{G}^s, \mathcal{C})$ 

\Statex  \hspace{1em}\textbf{Output:} Final causal graph $\mathcal{G}^s$, predicted\\ \hspace{1em}diagnosis $\mathcal{D}_{\text{pred}}$ and explanations $\mathcal{E}$.

\end{algorithmic}
\end{algorithm}

\section{Prompt Details}\label{apd:prompt}

In this section, we provide the prompt templates for knowledge retrieval agent, causal discovery agent, and decision-making agent, separately. 

\begin{tcolorbox}[colback=gray!10, colframe=gray!30, coltitle=black,  title=Knowledge retrieval agent - Prompt]
\small
\textcolor{cyan}{\textit{\# Knowledge retrieval:}}

\texttt{Generate a search query to retrieve the most relevant information from the knowledge database using \textcolor{cyan}{\{Diagnosis history\}} and \textcolor{cyan}{\{Candidate diseases\}}. The generated search query should take into account the characteristics of the knowledge database, as described by the provided \textcolor{cyan}{\{Meta-data\}}.}\newline \newline
\textcolor{cyan}{\textit{\# Reasoning in document:}}

\texttt{Summarize the \textcolor{cyan}{\{Document i\}}. The output summary should satisfy the following requirements:}

\texttt{Relevance: Include only information related to the patient’s \textcolor{cyan}{\{Diagnosis history\}} and \textcolor{cyan}{\{Candidate diseases\}}.}

\texttt{Conciseness: Remove redundant and unnecessary details while maintaining key insights.}

\texttt{Clarity: Ensure the summary is well-structured and easy to understand.}

\end{tcolorbox}

\begin{tcolorbox}[colback=gray!10, colframe=gray!30, coltitle=black,  title=Causal discovery agent - Prompt]
\small
\textcolor{cyan}{\textit{\# Hypothesis Generation:}}

\texttt{Generate a Directed Acyclic Graph (DAG) to represent the causal relationships between the given set of \textcolor{cyan}{\{Disease names\}}. Use the provided \textcolor{cyan}{\{Summary\}}, along with contextual knowledge and reasoning, to infer causality. The output should be in JSON format.}\newline \newline
\textcolor{cyan}{\textit{\# Hypothesis Amendment:}}

\texttt{Adjust the causal graph based on the current and previous versions stored in \textcolor{cyan}{\{Memory\}}, along with their fitting scores. Consider the following questions:}

\texttt{Are there any links that should be added?}

\texttt{Should any existing links be removed?}

\texttt{Should any directions be reversed?}

\texttt{Generate a revised causal graph and output it in a valid JSON format.}

\end{tcolorbox}

\begin{tcolorbox}[colback=gray!10, colframe=gray!30, coltitle=black,  title=Decision-making agent - Prompt]
\small
\texttt{Predict a list of diseases the patient may be diagnosed with in the future based on:}

\texttt{Patient summary and disease information: \textcolor{cyan}{\{Summary\}}}

\texttt{Causal DAG of disease relationships: \textcolor{cyan}{\{DAG.json\}}}

\texttt{Optional clinician comment: \textcolor{cyan}{\{Clinician comment\}}}

\texttt{Output format:}

\texttt{A JSON list of predicted ICD-9 codes.}

\texttt{A detailed explanation of the reasoning process.}

\texttt{Separate the two parts using the special token <SEP>.}

\end{tcolorbox}

\end{document}